\documentclass[runningheads]{llncs}
\usepackage{url}

\usepackage{todonotes}
\usepackage{graphicx}

\begin{document}
\title{Transferring façade labels between point clouds with semantic octrees while considering change detection}
\titlerunning{Transferring façade labels}
\author{Sophia Schwarz\textsuperscript{1 }, Tanja Pilz\textsuperscript{1 }, Olaf Wysocki\textsuperscript{1 }, Ludwig Hoegner\textsuperscript{1,2 }, Uwe Stilla\textsuperscript{1 }}
\authorrunning{Schwarz S., et al.,}
 \institute{Photogrammetry and Remote Sensing, TUM School of Engineering and Design, Technical University of Munich (TUM),\\ Munich, Germany - (sophiamaria.schwarz, tanja.pilz, olaf.wysocki, ludwig.hoegner, stilla)@tum.de \and Department of Geoinformatics, University of Applied Science (HM), Munich, Germany - ludwig.hoegner@hm.edu\\}
\maketitle              

\begin{abstract}

Point clouds and high-resolution 3D data have become increasingly important in various fields, including surveying, construction, and virtual reality. 
However, simply having this data is not enough; to extract useful information, semantic labeling is crucial. 
In this context, we propose a method to transfer annotations from a labeled to an unlabeled point cloud using an octree structure. 
The structure also analyses changes between the point clouds. 
Our experiments confirm that our method effectively transfers annotations while addressing changes.
The primary contribution of this project is the development of the method for automatic label transfer between two different point clouds that represent the same real-world object. 
The proposed method can be of great importance for data-driven deep learning algorithms as it can also allow circumventing stochastic transfer learning by deterministic label transfer between datasets depicting the same objects.
%\olafworries{one or two sentences of potential applications/implications}

\end{abstract}

\section{Introduction}

Point cloud data has gained popularity in recent years for its ability to provide high-resolution 3D data of real-world scenes. 
Point clouds are used in a variety of applications, including aiding navigation of autonomous cars~\cite{wilbers2019localization}, preserving cultural heritage~\cite{grilli2020machine}, and reconstructing 3D building models~\cite{HAALA2010570}.
Semantically enriched point clouds are particularly interesting for these applications since they not only provide geometric information but also add contextual information about real-world objects.
Recent advancements in deep learning have proven effective for automatic semantic point cloud labeling~\cite{tumfacadePaper}, which assigns labels to individual points. 
However, the performance of neural networks depends on large, manually labeled ground-truth datasets that are tedious and time-consuming to acquire. 
Providing point clouds with accurate labels is so far only possible by manual annotation.
This creates further problems in lacking ground-truth data for other research topics as well.

To tackle this issue, we introduce a method for automatic façade-semantic transfer between labeled and unlabeled point clouds representing the same building. 
We propose to achieve this goal by a semantic octree data structure preceded by a co-registration using the Iterative Closest Point concept. 
Additionally, the changes between the two point clouds are identified by utilizing the octree to describe the occupancy of the octree leaves.

\section{Related Work}

Generalized Iterative Closest Point (GICP) is a variant of the Iterative Closest Point (ICP) algorithm frequently used for point cloud co-registration~\cite{xu2021towards,segal2009generalized}. 
The ICP algorithm iteratively aligns two point clouds by minimizing the distance between corresponding point pairs~\cite{segal2009generalized}.
It iterates until the distance threshold is satisfied or a maximum number of iterations is reached.
Instead of relying solely on the distance between points, GICP minimizes the distance between geometric primitives (e.g., planes) found in point clouds.
This proves especially useful in an urban environment where regular, primitive structures are prevalent~\cite{xu2021towards}.  

Another related topic is the approach of utilizing octrees, and their 3D voxel leaves for change detection~\cite{kharroubi2022three}.
For instance, such a concept is employed by Gehrung et al., who utilize ray tracing on a highly-efficient octree structure and Bayesian reasoning to remove dynamic objects from point clouds based on three states:~\textit{occupied},~\textit{empty}, and~\textit{unknown}~\cite{gehrung2017approach}

Gehrung et al. \cite{gehrung2018avoxel} also proposed a fast voxel-based indicator for change detection using low-resolution octrees. 
This approach involves dividing the point cloud data into octree nodes and computing a histogram of voxel attributes, such as point density. 
Changes between two point cloud datasets can then be detected by comparing the histograms of corresponding nodes. The approach was shown to be effective in detecting changes in large-scale urban areas with low-resolution point cloud data.~\cite{gehrung2018avoxel}

\section{Methodology}

\begin{figure}[htbp]
\centering
\includegraphics[width=1\textwidth]{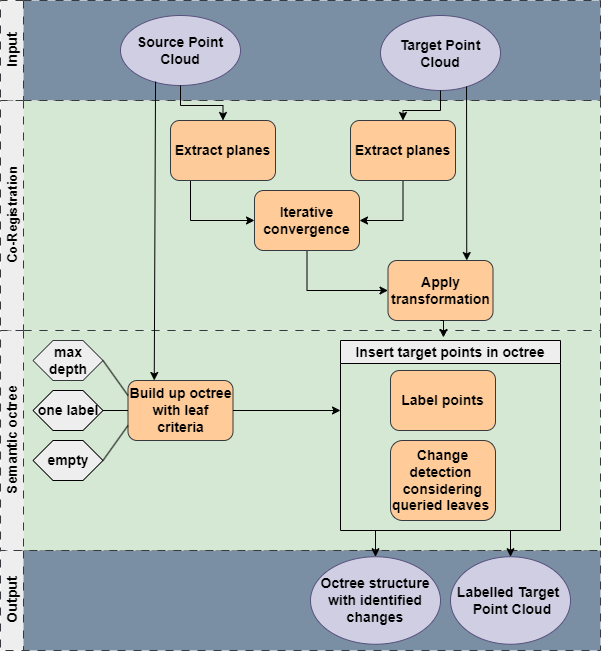}
\caption{Proposed workflow}
\label{fig:Workflow}
\end{figure}
The proposed method for the semantic label transfer between two urban point clouds uses an octree-based data structure that considers semantic information as a leaf criterion (Figure~\ref{fig:Workflow}). 
With the same octree-based data structure, coarse change detection is achieved.
Preceding this is a plane-based co-registration of the two point clouds that uses the Generalized Iterative Closed Point algorithm.
We developed the method based on work in the module Photogrammetry Project at Technical University Munich~\cite{dragicevic2023Transferring}. 
The implementation is available in the repository~\footnote{\url{https://github.com/SchwarzSophia/Transferring_urban_labels_between_pointclouds/}}. 

\subsection{Co-registration using Generalized Iterative Closest Point}

\begin{figure}[htbp]
\centering
\includegraphics[width=0.75\textwidth]{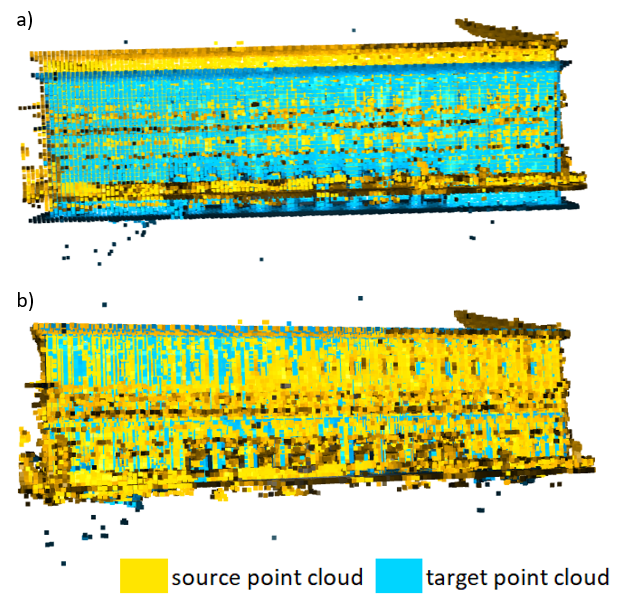}
\caption{Source and target point cloud a) before registration and b) after registration}
\label{fig:Comparison_GICP}
\end{figure}

As point clouds stemming from different campaigns can have global registration deviations (Figure~\ref{fig:Comparison_GICP}a), we first ensure that they are co-registered. 
We propose a method that leverages the planar-like structure of buildings by using plane-to-plane point cloud co-registration.
To additionally increase the robustness and simultaneously decrease computing time, a voxelizing precedes the GICP~\cite{Koide2021}. 
For this, we compute the estimated normals for the downsampled source and target points. 
Next, we calculate the registration using the two input point clouds, the initial transformation matrix, the maximal corresponding distance between a plane-pair $\lambda$, and the convergence criteria $\chi$ (Figure~\ref{fig:Workflow}).
After calculating the transformation matrix based on the downsampled points, we apply it to the complete dataset (Figure~\ref{fig:Comparison_GICP}b).
As presented in Figure \ref{fig:Comparison_GICP} b, the registered point clouds do not show the previously noted deviations.

\subsection{Semantic Octree}
The data structure of the semantic octree performs two tasks simultaneously: transferring labels and detecting changes. 

We introduce two attributes to the octree leaves, a semantic label and a label describing whether a region has been removed.
Considering a dynamic depth, the octree is set up only using the semantically labeled point cloud.
While building up the octree, a node is defined as a leaf when it meets one of three leaf-criteria~\textit{empty leaf},~\textit{one label} and~\textit{max depth}, which are checked in this order.

A leaf is considered as~\textit{empty leaf} if no points are contained in its volume.
We define an empty leaf as a \textit{new} label, indicating new points compared to the target cloud, whereas \textit{unchanged} are the matching areas.
In this case, the semantic label indicates an unknown class, which later describes newly captured façade points.
This is used to characterize added building parts in the change detection.

The criterion~\textit{one label} is fulfilled if all points in a node have the same semantic class. 
This class is defined as the semantic leaf label.
The last criterion to be checked is~\textit{max depth}, referring to the maximum octree depth a leaf might reach.
This prevents over-fitting of the data and, at the same time, shortens the computation time.
If the maximal depth is reached and the labels in the leaf container are heterogeneous, the prevalent label in the container is identified as the semantic leaf label. 
The~\textit{removed} label is set to true at this stage. 
The maximum depth is controlled by~\textit{max lat}, the largest permitted side length of the smallest leaf container:
Formula~\ref{depth} specifies the relationship between the lateral point cloud length, the maximal lateral length of the smallest octree leaf, and the depth of the octree.

\begin{equation}
\label{depth}
depth = \log_2\left(\frac{lateral \ point \ cloud\ length}{maxLat}\right)
\end{equation}

With these criteria, we design the data structure to adapt to the density of the point cloud and the complexity of the depicted building.
To label the points of the target point cloud, they are iteratively inserted into the octree. 
During this process, two determinations are made: the point is assigned the label of the leaf it falls into, and the~\textit{removed} label of the queried leaf is set to false.
Finally, the target point cloud is labeled, and the octree leaves represent a coarse change detection.

\section{Experiments}

\begin{figure}[htbp]
\centering
\includegraphics[width=1\textwidth]{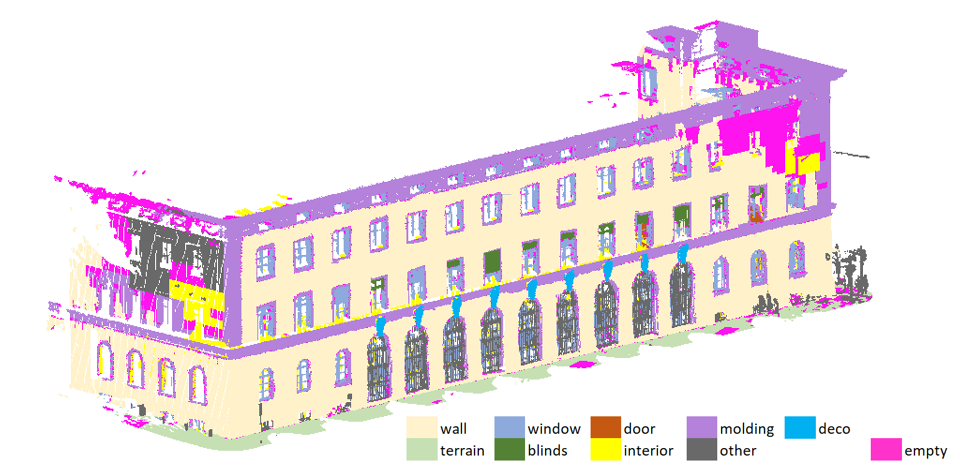}
\caption{MF point cloud with automatically transferred labels at a lateral length of 10 cm}
\label{fig:Ganz_10_Legende}
\end{figure}

We used the open TUM-FAÇADE~\cite{tumfacadePaper} dataset and the proprietary MF point cloud for the experiments.
The TUM-FAÇADE segmentation benchmark served as a labeled point cloud, whereas MF was the unlabeled one.
We manually labeled the MF point cloud to validate our method and evaluate our generated results.
We determined optimal values for the co-registration experimentally. 
We achieved the best performance with a voxel size $\nu$ of 10 cm and a maximal correspondence distance of $\lambda$ = 10 m. 
For the convergence criteria, we set the number of maximal iterations to 50 and both the relative fitness and relative root mean square error to 1 µm.
The parameters for the co-registration are dependent on the datasets in use.
However, we can assume that similar values are reasonable for all point clouds representing urban buildings. 

The datasets were pre-processed with FME (Feature Manipulation Engine) and Python using the Open3D library~\cite{Zhou2018Open3D} to ensure uniformity in the coordinate system across all input datasets. 
Additionally, statistical outlier removal techniques were applied to the point clouds, again utilizing Open3D functions~\cite{Zhou2018Open3D}.
The statistical outlier removal process involved evaluating the neighborhood density and the statistical distribution of densities for each point. 
A threshold was established to allow for permissible deviations from the mean density. 
As indicated by the threshold, points within sparser neighborhoods were subsequently removed from the analysis.
Our study employed a threshold of 1.7 $\sigma$, considering 20 neighboring points for density calculations and outlier detection. 
This specific parameter configuration allowed for effectively removing points exhibiting significantly deviant density values.
By implementing these pre-processing steps, including coordinate system transformation and statistical outlier removal, we ensured a consistent and refined dataset for subsequent analysis and interpretation.

\subsection{Quantitative Analysis}

\begin{table}[htbp]
  \centering
  \includegraphics[width=\textwidth]{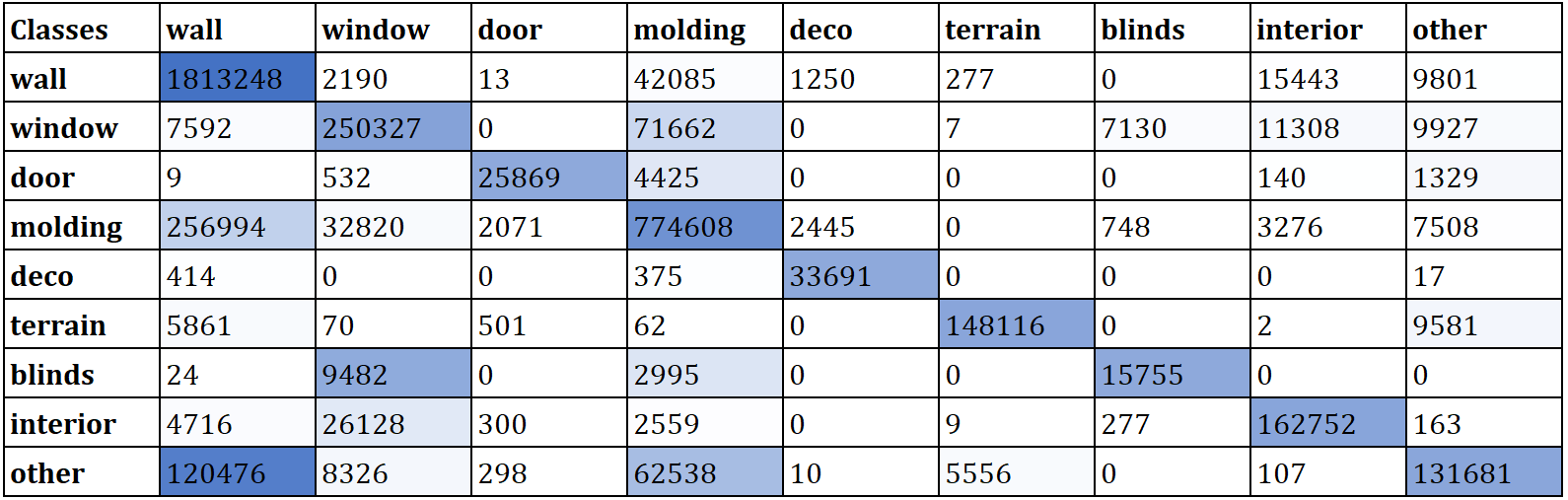}
  \caption{Confusion matrix of the example building calculated with a maximal lateral length of 10 cm}
  \label{tab:Confusionsmatrix}
\end{table}

For evaluation, we compared the computed labels to the labels of the manually classified MF.
For this, we calculated a confusion matrix for different octree depths, the overall accuracy, and the Cohen Kappa coefficient. 

\begin{table}
     \centering
     \begin{tabular}{|l|l|l|}
     \hline
     \bfseries{lateral lengths} & \bfseries{overall accuracy} & \bfseries{kappa coefficient}\\
     \hline
     \bfseries{5 cm} &0.818 &0.725 \\
     \hline
     \bfseries{10 cm} &0.817 &0.723 \\
     \hline
     \bfseries{20 cm} &0.815 &0.722 \\
     \hline
     \end{tabular}
     \caption{Overall accuracy and kappa coefficient for three different lateral lengths}
     \label{tab_accuracy}
 \end{table}

We observe an overall accuracy of about 81\% and a kappa coefficient of around 0.72 (Table~\ref{tab_accuracy}). 
The three different lateral lengths shown have deviations in the third decimal, which is an insignificant difference. 
These results indicate a good matching within the classification.

The TUM-FAÇADE point cloud has a higher point density than the manually labeled cloud (reflected by Table~\ref{tab_numberpoints}). 
This deviation can be one of the reasons for inaccuracies in the method.

The identified changes are depicted in Figure~\ref{fig:changes_10cm_Legende}.
The new points make up around 7.9 \%. 

\begin{table}
\centering
\begin{tabular}{|l|l|}
\hline
\bfseries{Point Cloud} & \bfseries{Total number of points}\\
\hline
TUM-FAÇADE & 35 837 MLN\\
\hline
MF & 4 461 MLN\\
\hline
\end{tabular}
\caption{Total number of points in TUM-FAÇADE and manually labeled point cloud}
\label{tab_numberpoints}
\end{table}

\subsection{Qualitative Analysis}
\begin{figure}[htbp]
\centering
\includegraphics[width=1\textwidth]{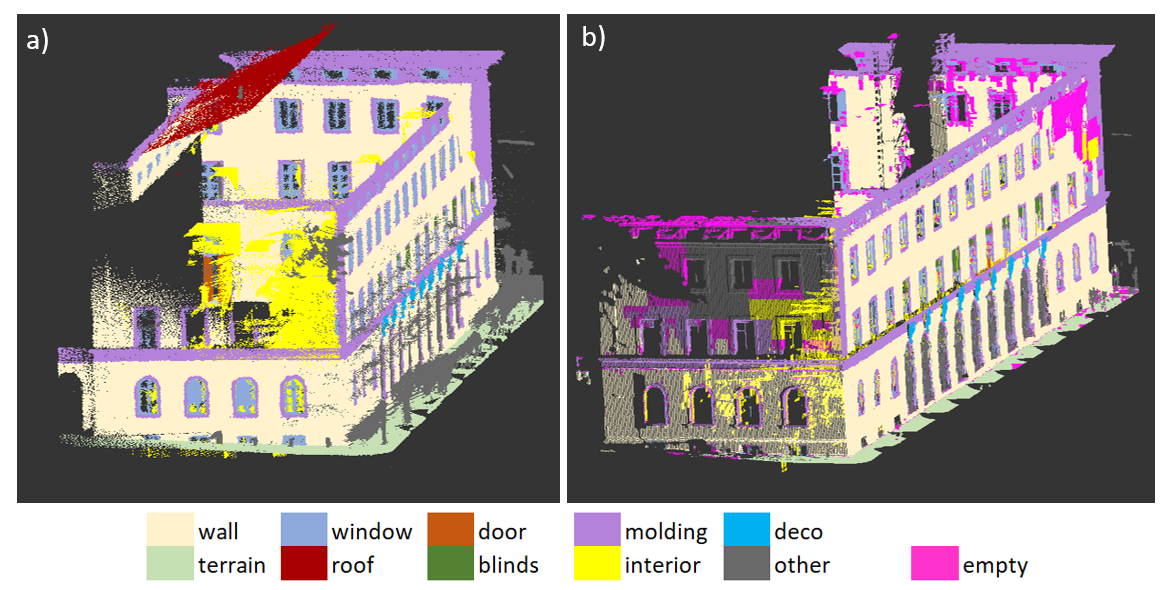}
\caption{South corner of the example building of (a) the source point cloud TUM-FAÇADE~\cite{tumfacadePaper} and (b) the resulting point cloud}
\label{fig:Comparison_gray_hole}
\end{figure}
 
\begin{figure}[htbp]
\centering
\includegraphics[width=1\textwidth]{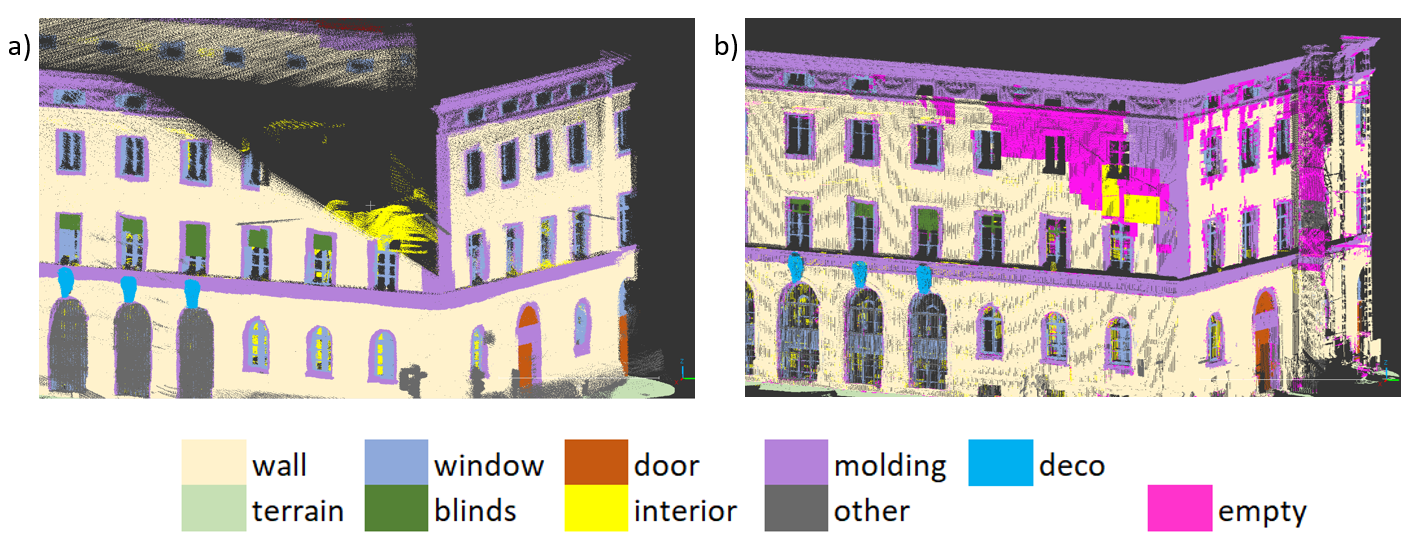}
\caption{North-east corner of the example building of (a) the source TUM-FAÇADE~\cite{tumfacadePaper} and (b) target point cloud MF}
\label{fig:Comparison_Corner}
\end{figure}

Figure~\ref{fig:Ganz_10_Legende} provides a visualization of the result of our proposed method with the point cloud changes.
As indicated in the results of the quantitative analysis, the method works well.
The new class~\textit{empty}, generated by the method, is colorized in pink.

During our visual evaluation, we detected several inaccuracies in the complex features of the point cloud, e.g., the windows or the blinds as well in the areas that do not show holes in the source point cloud.
The comparison in Figure~\ref{fig:Comparison_Corner} illustrates the method's influence on holes within the source point cloud. 
In areas where the point cloud exhibited sufficient density, accurate label transfer was achieved, ensuring reliable results. 
However, in regions where the point cloud lacked density, inaccuracies surfaced. 
Specifically, rather than uniformly designating the entire hole with class 18, a combination of diverse classes became evident within the hole. 
For instance, the distinct features of the building's corner molding appeared blended into the hole's context, contributing to the observed inaccuracies.

The most noticeable error occurred at a hole on the south side of the building. 
Instead of assigning the class~\textit{empty}, most of the points in the concerned area were assigned the class~\textit{other}.

\begin{figure}[htbp]
\centering
\includegraphics[width=1\textwidth]{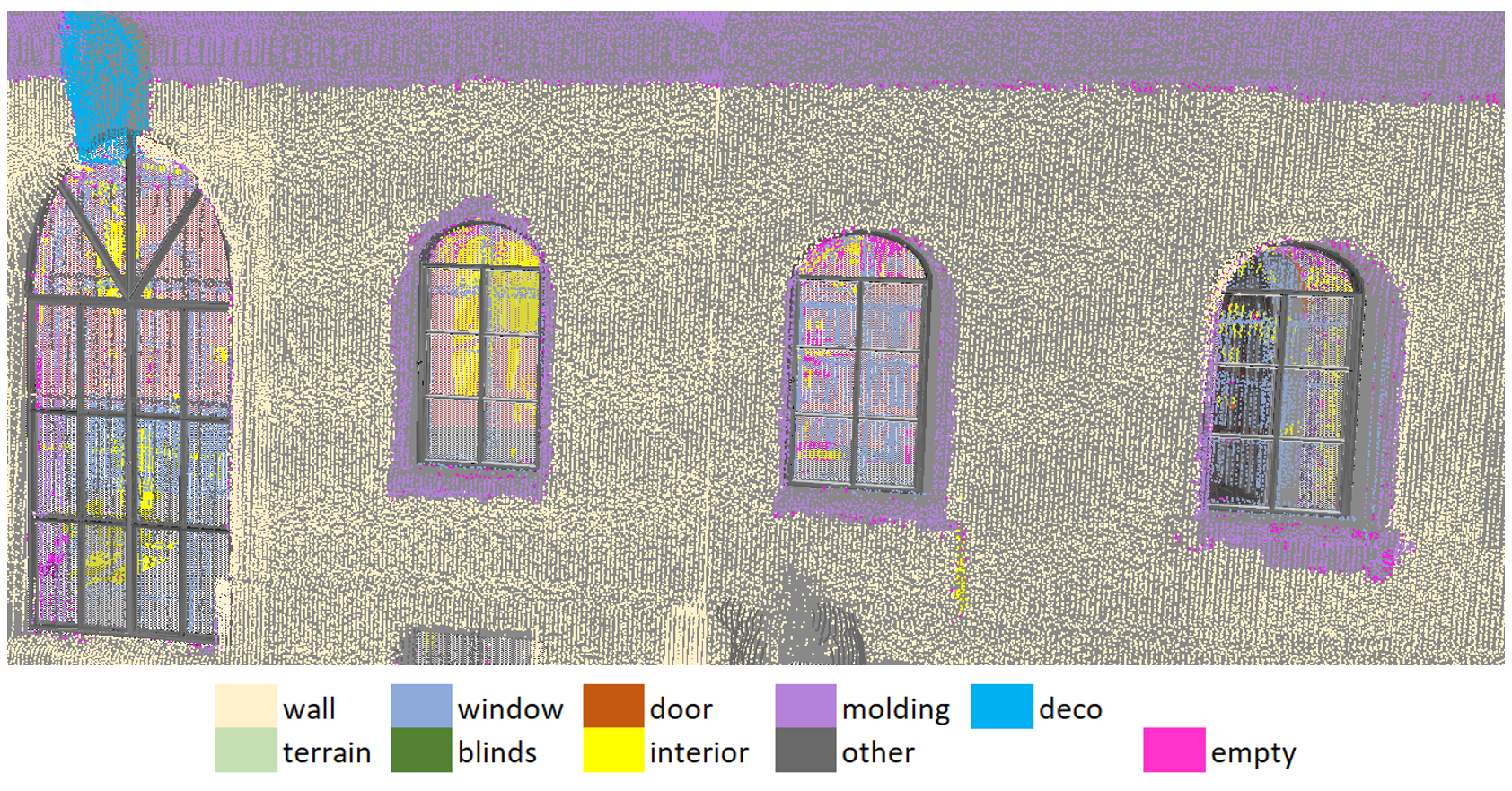}
\caption{Close-up of exemplary windows }
\label{fig:window_closeup}
\end{figure}

With a more detailed view, the quality of the labeled point cloud can be better assessed.  
Minor inaccuracies around complex structures are evident in the close-up of window structures, Figure~\ref{fig:window_closeup}. 
While the wall was consistently labeled and the location of the window contours are transferred correctly, they are not precisely defined at the edges.

\begin{figure}[htbp]
\centering
\includegraphics[width=1\textwidth]{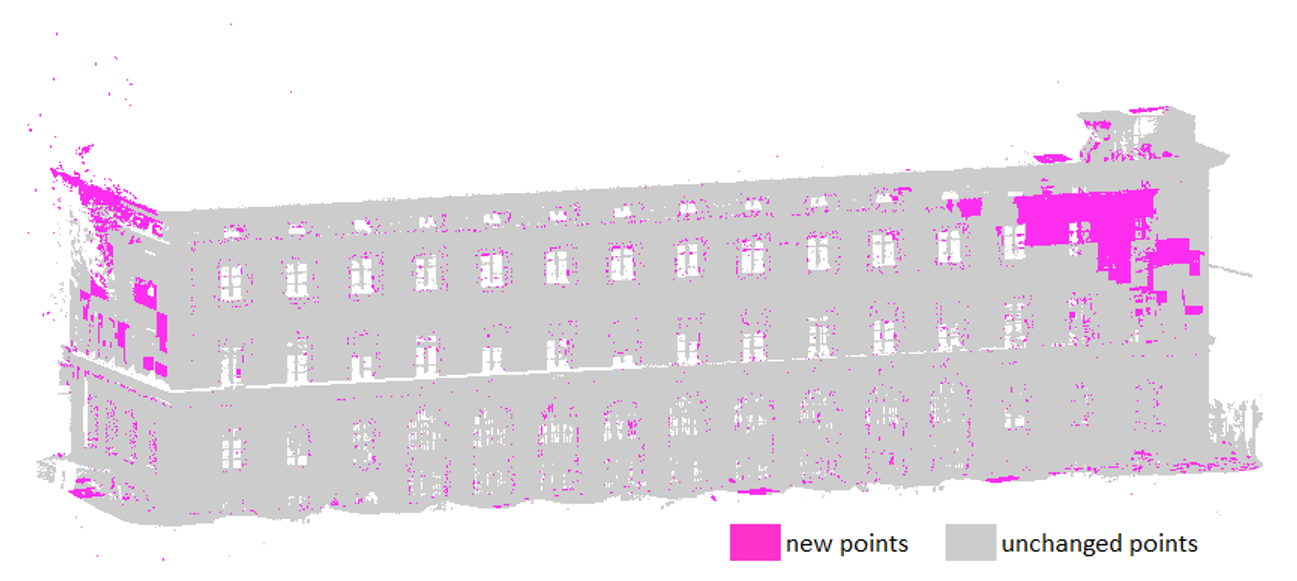}
\caption{Change detection with a lateral length of 10 cm}
\label{fig:changes_10cm_Legende}
\end{figure}

In Figure~\ref{fig:changes_10cm_Legende}, we see the algorithm's performance with changes in the point clouds by comparing the~\textit{new} and the~\textit{unchanged} points.
Our method detects rough changes in the building effectively.
Fine changes, however, that do not change the geometric structure of the point cloud are not registered. 

\subsection{Discussion}

Experiments revealed promising results for transferring labels between clouds captured from the same building façade using the proposed method.
For the majority of the points, the method assigned the correct labels.
Subsequently, we discuss the reasons for the remaining inaccuracies.

In our analysis, several inaccuracies occurred in the representation of complex features, such as windows or blinds (see Figure~\ref{fig:Comparison_Corner}b). 
These deviations can be attributed to the parameter~\textit{max depth} limitations.
If structures are smaller than this parameter, they cannot be represented accurately.

Further inaccuracies arose with round façade elements. 
They were not transferred precisely because the algorithm works with cubic volumes. 
Therefore, the rounding can only be approximated, the accuracy of which depends on the~\textit{max depth} parameter.
Nonetheless, defining the parameter is necessary to avoid over-fitting the octree to the source point cloud, which would cause too many small leaves with the label~\textit{new}, leading to wrongly labeled~\textit{new} points. 
We found an optimal~\textit{max depth} of 10 centimeters lateral length for our data in experiments.

Additionally, different point densities in point clouds (as shown in Table~\ref{tab_numberpoints}) pose a significant challenge. 
If the source point cloud has a lower point density than the target point cloud, the achievable accuracies are limited from the start. 
Since the octree structure considers the local density of the points, large cubes with few points may not be further subdivided.
 
This can lead to new features being incorrectly assigned a class, as illustrated in more detail in Figure~\ref{fig:Comparison_gray_hole}.
We had larger holes in the source point cloud in our example building. 
Our proposed method handles this well.
As shown in Figure~\ref{fig:Comparison_Corner}, all added points are labeled as~\textit{empty}.
Nevertheless, there can be inaccuracies at the edge of the holes. 
Again, the volume is not further divided because only a few homogeneous points fall into the edge area.
At this point, the limitations of change detection become apparent. 
Rough changes are well detected by the method. 
However, subtle changes that do not alter the structure of the building are not registered. 
The reason for this is the limitation of the method, as it cannot deal with the change of objects in the same place, for instance, recording open versus closed window blinds. 
In both cases, there are points in the cube, so no empty areas are detected. 
The points are not identified as changed but are assigned the wrong label.

Lastly, we acknowledge that some mislabeled points can stem from discrepancies in the manually labeled test data sets.
These are due to different interpretations of points, leading to varied classifications or simply small errors in labeling. 
While these biases do not necessarily distort the test results gravely, they can still introduce inconsistencies in the final results.

\section{Conclusion}

In this paper, we present a method that utilizes generalized Iterative Closed Point algorithm for point cloud registration and an octree map to transfer annotations from a labeled point cloud to an unlabeled one. 
Additionally, we detect changes by utilizing the semantic octree.
Our experiments corroborate the effectiveness of the proposed method in accurately transferring labels between point clouds (approx. 82 \%) with similar geometric and semantic structures.
However, we observe a few inaccuracies in the method, especially when looking at complex or round features.

As a next step, the edge blurring could be improved by introducing a minimal octree depth and dividing the nodes into containers according to their local density. 
Lastly, the method can be compared to other state-of-the-art methods for semantic information transfer to determine its relative strengths and weaknesses. 
Further research can help to refine and improve this for practical use.
This method can be of great importance for data-driven deep learning algorithms as it can also allow circumventing stochastic transfer learning by deterministic label transfer between datasets depicting the same objects.

\section{Acknowledgements}

This work was supported by the Bavarian State Ministry for Economic Affairs, Regional Development, and Energy within the framework of the IuK Bayern project \textit{MoFa3D - Mobile Erfassung von Fassaden mittels 3D Punktwolken}, Grant No.\ IUK643/001.
Moreover, the work was conducted within the framework of the Leonhard Obermeyer Center at the Technical University of Munich (TUM).
We gratefully acknowledge the Geoinformatics team at the TUM for the valuable insights and for providing the CityGML datasets.
This method was developed in the module Photogrammetry Project of the Master's study program Geodesy and Geoinformation at the TUM. 
We thankfully acknowledge the supervisor team for their conductive insights and support.
%
% ---- Bibliography ----

%
\bibliographystyle{splncs04}
\bibliography{bibliography}

\end{document}